\documentclass[11pt]{article}

\usepackage[utf8]{inputenc}
\usepackage[T1]{fontenc}
\usepackage[margin=1in]{geometry}
\usepackage{amsmath,amssymb}
\usepackage{booktabs}
\usepackage{listings}
\usepackage{xcolor}
\usepackage[expansion=false]{microtype}
\usepackage{hyperref}
\usepackage{enumitem}
\usepackage{caption}
\usepackage{graphicx}
\hypersetup{colorlinks=true,linkcolor=blue!50!black,citecolor=blue!50!black,urlcolor=blue!50!black}

\definecolor{codebg}{gray}{0.96}
\title{\textbf{The Log is the Agent}\thanks{Open source (Apache-2.0): \texttt{https://github.com/yoheinakajima/activegraph}. Install with \texttt{pip install activegraph} and reproduce the worked example with \texttt{activegraph quickstart}. Documentation: \texttt{https://docs.activegraph.ai}.}\\[0.3em]
\large Event-Sourced Reactive Graphs for Auditable, Forkable Agentic Systems}

\author{Yohei Nakajima\\
\small Untapped Capital \\
\small \texttt{activegraph.ai}}

\date{\today}

\begin{document}
\maketitle

\begin{abstract}
\noindent
Most agent frameworks are built \emph{around} the language model: a conversation
loop comes first, then tools, then rules, and finally a logging layer bolted on
for observability, with state persisted as retrievable ``memory.'' We describe
ActiveGraph, a runtime that inverts this arrangement. The append-only event log
is the source of truth; the working graph is a deterministic projection of that
log; and behaviors---ordinary functions, classes, LLM-backed routines, or logic
attached to typed edges---react to changes in the graph and emit new events. No
component instructs another; coordination happens entirely through the shared
graph. This single design decision yields three properties that
retrieval-and-summarization memory systems do not provide: deterministic
\emph{replay} of any run from its log, cheap \emph{forking} that branches a run
at any event without re-executing the shared prefix, and end-to-end
\emph{lineage} from a high-level goal down to the individual model call that
produced each artifact. We present the architecture, a determinism contract that
makes replay sound, and a worked diligence example whose full causal structure
is reconstructable from the log alone. We discuss---without claiming to
demonstrate---why this substrate is unusually well suited to self-improving
agents, and how it extends the BabyAGI lineage and prior graph-memory research.
\end{abstract}

\section{Introduction}

The dominant pattern for building LLM agents grows by accretion. We begin with a
chat loop, because conversation is the interface the model was trained for. We
add tools when the model needs to act on the world. We add rules and guardrails
when the behavior drifts. We add logging because, in production, we need to know
what happened. And we store some compressed form of the interaction---summaries,
embeddings, a vector index---so the agent can ``remember'' across turns. In this
arrangement the log is a \emph{byproduct}: an audit artifact written alongside
the real computation, never the substrate of it.

This paper describes a runtime built on the opposite assumption. What if the log
were not a byproduct but the agent itself? Concretely: what if the rules an agent
follows (and the changes to those rules), the tools it is given and the calls it
makes, and the content it produces were all the same kind of thing---events in a
single append-only log---and the agent's behavior were nothing more than a set of
reactions that fire when that log grows and write new events back into it?

ActiveGraph is that runtime. The event log is the source of truth. The graph that
behaviors read and write is a projection of the log, recomputed by a
\emph{replay} operation that is guaranteed deterministic. Behaviors subscribe to
patterns over events and graph shape; when a matching change occurs, the behavior
fires, possibly calls a model or a tool, and emits the resulting facts as new
events. There is no orchestrator threading state between steps, and there are no
workflows in the usual sense---only a population of rule-like behaviors that
sometimes chain because one behavior's output is another's trigger.

This is a direct descendant of BabyAGI~\cite{babyagi}, which expressed an agent as
a \texttt{while}-loop over a global task list: execute the current task,
summarize it against the objective, generate follow-up tasks. ActiveGraph
re-expresses that same loop as reactive behaviors over a shared graph, and in
doing so converts the agent's transient state into a durable, inspectable,
replayable artifact.

\paragraph{Contributions.} This is a systems paper; its contributions are
architectural, not empirical claims about task performance:

\begin{enumerate}[leftmargin=1.4em,itemsep=2pt]
  \item \textbf{An event-sourced agent model} in which graph state is a
  deterministic fold over an append-only event log, and all agent action is
  expressed as behaviors that react to graph changes and emit events
  (\S\ref{sec:motivation}, \S\ref{sec:architecture}).
  \item \textbf{A determinism contract and replay mechanism} that makes any run
  byte-reproducible from its log, including a content-addressed cache that records
  model and tool responses so replay performs no new model calls
  (\S\ref{sec:replay}).
  \item \textbf{A forking and structural-diff primitive} that branches a run at
  any event and answers counterfactual ``what if I had done X differently''
  questions without re-executing the shared prefix (\S\ref{sec:forking}).
  \item \textbf{A worked, fully reproducible example} (an investment-diligence
  pack) in which the entire causal chain from goal to memo is recoverable from
  the event log, demonstrating lineage as a first-class deliverable
  (\S\ref{sec:worked}).
\end{enumerate}

We are careful about what we do \emph{not} claim. We do not report that
ActiveGraph improves task accuracy over any baseline; the contribution is the
substrate and its guarantees. We discuss self-improving agents in
\S\ref{sec:selfimprove} strictly as an affordance the architecture enables, not
as a result we evaluate.

\section{Motivation: built around the model versus built on the log}
\label{sec:motivation}

Consider what an agent's state actually consists of. There is the objective. There
are the rules it operates under, which may themselves change mid-run. There are
the tools available and the record of which were called with what arguments. There
is the conversation or reasoning trace. And there is whatever the agent has
produced. In the conventional architecture these live in different places: the
objective in a prompt, the rules in code or a system message, the tool calls in
a framework's internal state, the trace in a transcript, the outputs in a
database, and a lossy projection of all of it in a ``memory'' store queried by
similarity.

ActiveGraph collapses these into one substrate. Every one of them is an event.
A rule change is an event. A tool call and its response are events. A produced
claim is an event that creates an object, and the link from that claim to the
evidence supporting it is another event that creates a relation. Because they are
all events in one ordered log, the questions that are awkward or impossible in the
conventional architecture become trivial: \emph{Why is this fact in the agent's
working set? What did the agent believe before it changed rule R? What would have
happened had it taken the other branch at step 42?} Each of these is a query over,
or a re-projection of, the log.

It is tempting to motivate this by analogy to human cognition---we, too, are
arguably shaped less by a reasoning engine than by an accumulated history of
experiences and the beliefs they produced. We note the analogy only in passing
and rest no argument on it; the case for the architecture is mechanical, not
cognitive, and is made in the sections that follow.

The borrowed ideas are not new in isolation. Event sourcing and
command-query responsibility segregation are established patterns in data
systems, and reactive recomputation of derived state from a changing input is the
core of incremental dataflow and of spreadsheet engines. The contribution here is
the recombination: applying an event-sourced, reactive substrate to the specific
problem of long-running agentic systems, where the payoff---auditable lineage,
deterministic replay, and cheap counterfactual forks over computations that
include nondeterministic model calls---is both non-obvious and, we argue,
unusually valuable.

\section{Architecture}
\label{sec:architecture}

\begin{figure}[t]
\centering
\includegraphics[width=\textwidth]{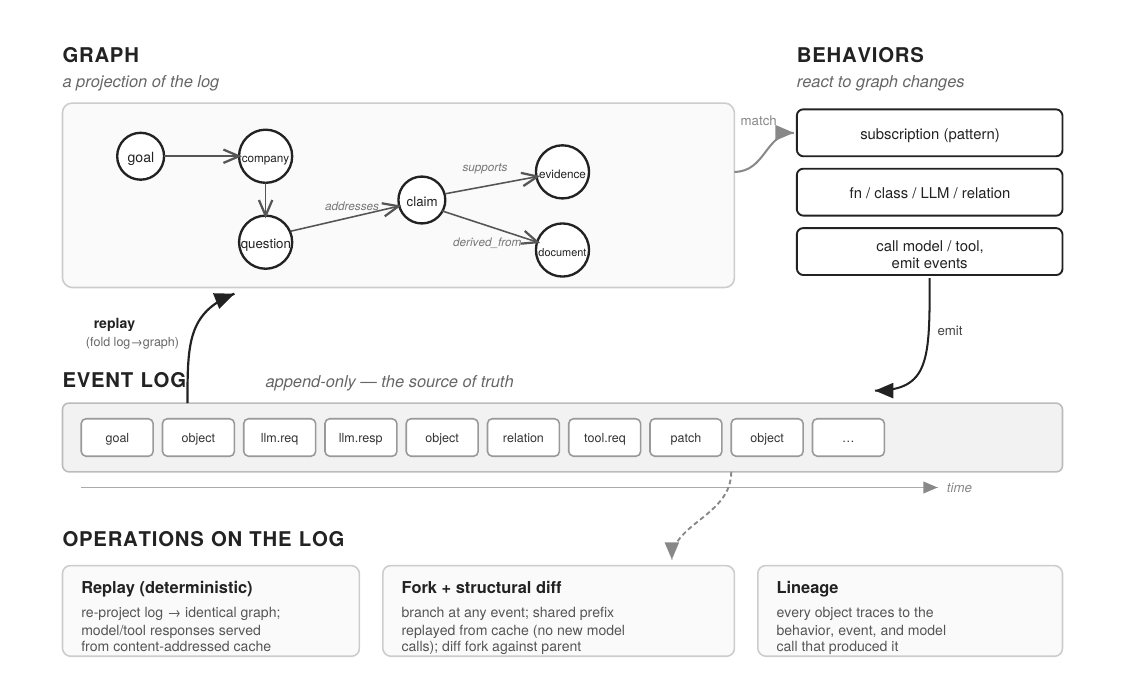}
\caption{The ActiveGraph runtime as a cycle. The append-only event log (bottom) is
the source of truth. \emph{Replay} folds the log into the graph (a projection of
typed objects and relations). Behaviors subscribe to graph-shape patterns, react to
matching changes by calling models or tools, and emit new events back into the log.
Operations on the log---deterministic replay, fork-and-diff, and lineage---follow
from the log being primary.}
\label{fig:arch}
\end{figure}

\paragraph{The graph is a projection of the log.}
The runtime holds an append-only sequence of events. Graph state---a typed set of
objects and relations---is never mutated directly by external code; it is computed
by folding the event log forward. Loading a run from a store, forking it, or
validating it all invoke the same underlying \emph{replay} operation that rebuilds
the graph from its events. Two replays of the same log produce deterministically
identical state.

\paragraph{Events.}
Every event carries an id, a type, a payload, the actor that produced it, an
optional \texttt{caused\_by} pointer to the event that triggered it, and a
timestamp. Listing~\ref{lst:events} shows the opening events of a real run: a pack
loads, a goal is created by the user, a behavior starts in response, and an object
is created carrying a \texttt{provenance} block that names the behavior that
created it and the event that caused it.

\begin{lstlisting}[language=,caption={Opening events of a real run (abridged from the captured 671-event log). Each object's \texttt{provenance} records the behavior and causing event; nothing is created without a traceable origin.},label={lst:events}]
{
  "id": "evt_001",
  "type": "pack.loaded",
  "actor": "runtime",
  "caused_by": null,
  "payload": {
    "name": "diligence",
    "version": "0.1.0",
    "object_types": [
      "company",
      "document",
      "question",
      "claim",
      "..."
    ],
    "behaviors": [
      "diligence.company_planner",
      "diligence.question_generator",
      "..."
    ]
  }
}
{
  "id": "evt_002",
  "type": "goal.created",
  "actor": "user",
  "caused_by": null,
  "payload": {
    "goal": "Diligence: Northwind Robotics"
  }
}
{
  "id": "evt_003",
  "type": "behavior.started",
  "actor": "runtime",
  "caused_by": "evt_002",
  "payload": {
    "behavior": "diligence.company_planner",
    "event_id": "evt_002"
  }
}
{
  "id": "evt_004",
  "type": "object.created",
  "actor": "diligence.company_planner",
  "caused_by": "evt_002",
  "payload": {
    "object": {
      "id": "company#1",
      "type": "company",
      "data": {
        "name": "Northwind Robotics"
      },
      "provenance": {
        "created_by": "diligence.company_planner",
        "caused_by_event": "evt_002"
      }
    }
  }
}
{
  "id": "evt_005",
  "type": "behavior.completed",
  "actor": "runtime",
  "caused_by": "evt_002",
  "payload": {
    "behavior": "diligence.company_planner",
    "event_id": "evt_002"
  }
}
\end{lstlisting}

\paragraph{Behaviors.}
A behavior is a reaction. It declares a subscription---an event type plus an
optional predicate and a graph-shape pattern expressed in a Cypher
subset---and a body. When a matching change lands on the graph, the runtime fires
the body, supplying the triggering event, a view of the graph, and a context
handle. The body may create objects and relations, apply patches to existing
objects, call tools, or call a model; whatever it does is recorded as events.
Bodies come in four forms: a plain function, a class (for behaviors that carry
configuration), an LLM-backed routine (whose request and response are themselves
logged events), and a \emph{relation-behavior}---logic attached to a typed edge,
so that the act of relating two objects can itself carry computation.

\paragraph{Why a graph, and not just a log.}
Event sourcing alone would give a log and a flat projection of current values; the
graph earns its place by changing what behaviors can \emph{subscribe to} and what
operations are cheap. Three things depend on it specifically. First, subscriptions
are graph-shape patterns, not just event-type filters: a behavior can fire on
``a \texttt{claim} that \texttt{addresses} an unanswered \texttt{question},'' a
predicate over topology that a flat event stream cannot express without the
consumer rebuilding the graph itself. Second, relation-behaviors make typed edges
first-class carriers of logic, so computation attaches to the \emph{relationship}
between objects, not only to the objects---there is no natural place for this in a
keyed key-value projection. Third, the structural diff of \S\ref{sec:forking} is a
diff over graph topology (which objects, relations, and patches differ between two
runs), which is well-defined precisely because the projection is a graph rather
than an opaque blob. The log is what makes state reproducible; the graph is what
makes reactivity and comparison expressible. Neither alone is sufficient.
Table~\ref{tab:compare} summarizes how these properties distinguish the
log-primary design from conventional agent loops and from memory-layer systems.

\begin{table}[ht]
\centering
\small
\caption{Where the architectures differ. ActiveGraph's column follows from treating
the event log as primary rather than as a byproduct. ``Partial'' marks properties
a system offers in a limited or non-guaranteed form.}
\label{tab:compare}
\begin{tabular}{lccc}
\toprule
Property & Conventional & Memory-layer & ActiveGraph \\
 & agent loop & systems & \\
\midrule
Persistent state across sessions & partial & yes & yes \\
Provenance: why is this in context? & no & partial & total \\
Deterministic state reconstruction & no & no & yes \\
Replay of a full run from its history & no & no & exact \\
Fork a run at an arbitrary point & no & no & yes \\
Fork cost on the shared prefix & re-execute & re-execute & cached \\
Structural diff between two runs & no & no & yes \\
\bottomrule
\end{tabular}
\end{table}

\paragraph{No workflow---but not no coordination.}
There is no top-level script sequencing the steps of a run. The diligence example
in \S\ref{sec:worked} produces 93 objects and 76 relations across three companies
without a single line of orchestration code: a planner behavior reacts to a goal
by creating a company; a question-generator reacts to the company by emitting
research questions; a researcher reacts to each question; and so on. The control
flow is an emergent consequence of which events match which subscriptions. We are
careful not to overstate this: coordination has not vanished, it has moved. What
was explicit orchestration in a workflow engine becomes implicit coordination
through the shared graph and its subscriptions. The claim is not that coordination
disappears but that making it implicit and data-driven---rather than encoding it
in a control-flow script---is what makes runs uniformly replayable, forkable, and
inspectable, because every coordinating decision is itself an event rather than a
position in some external program counter.

\paragraph{The determinism contract.}
Because state is a fold over the log, replay can only be sound if behavior bodies
are deterministic functions of their inputs. The runtime therefore imposes a
contract: a behavior body must not read \texttt{random}, wall-clock time, or fresh
UUIDs directly (it obtains these from the event, which carries a recorded
timestamp, or from the runtime's deterministic id generator); it must not perform
I/O outside the framework's tool and model primitives; and it must not depend on
mutable global state that changes across fires. The contract is not statically
enforced. A behavior that violates it runs correctly the first time and is caught
later, at replay or fork, as a divergence error pinned to the first event that
fails to reproduce. The next section explains why that is sufficient.

The apparent exception is the case that matters most: an LLM-backed behavior,
which is what makes ActiveGraph an agent rather than a workflow engine. A model
call is not a deterministic function of its inputs, so such a behavior does not
satisfy the contract at \emph{first execution}. The framework does not require it
to. The call goes out live, and its response is recorded as an event; the contract
applies to \emph{replay}, where that recorded response is served from cache (\S\ref{sec:replay})
and the behavior consequently reproduces exactly. Determinism is thus a property
of re-projecting a log that already exists, never a claim that running the agent
is reproducible. This is a deliberate division of labor: behaviors may freely call
models and tools at execution time, and the recording layer is what makes the
resulting run replayable after the fact.

\section{Replay and determinism}
\label{sec:replay}

Replay is the operation that reconstructs graph state from the log. It is the
foundation for everything that follows: forking replays a shared prefix, audit
replays the whole run, and validation replays in a strict mode that proves
reproducibility. The runtime guarantees that, given a log, replay is deterministic
and produces identical state on every run.

\paragraph{The honest problem: models are not deterministic.}
A behavior that calls a language model is not, in general, a deterministic function
of its inputs---the same prompt can yield different completions. ActiveGraph does
not pretend otherwise. Instead it makes replay deterministic by \emph{recording}
model and tool responses rather than by assuming they are reproducible. The cache
is content-addressed: a model response is keyed on a hash of the entire request
(system message, user messages, model identifier, tool definitions, and output
schema); a tool response is keyed on the tool name and a deterministic hash of its
arguments. During replay, when a behavior re-fires with a request whose hash
matches a recorded one, the cached response is served from the log's
\texttt{llm.responded} or \texttt{tool.responded} event. No new model call is made.
The cost is disk space bounded by the size of the run.

This is visible directly in the log. Listing~\ref{lst:llm} shows a real
\texttt{llm.requested} event: the request is normalized, hashed, and marked
\texttt{deterministic} with \texttt{temperature} pinned to zero, and the
corresponding response event carries the prompt hash that indexes it for replay.

\begin{lstlisting}[language=,caption={A real \texttt{llm.requested} event (abridged). The request is content-hashed and the response is recorded under that hash, so a later replay or fork serves it from cache rather than calling the model again.},label={lst:llm}]
{
  "id": "evt_007",
  "type": "llm.requested",
  "caused_by": "evt_004",
  "payload": {
    "behavior": "diligence.question_generator",
    "model": "claude-sonnet-4-5",
    "prompt_hash": "c05af4931bdd7bc4...",
    "deterministic": true,
    "cache_hit": false,
    "prompt": {
      "temperature": 0.0,
      "top_p": 1.0,
      "deterministic": true,
      "output_schema_name": "QuestionList"
    },
    "estimated_cost_usd": "0.001"
  }
}
\end{lstlisting}

\paragraph{Strict versus permissive replay.}
Replay runs in one of two modes. In \emph{permissive} mode---the default when
loading a run---events are re-emitted from the log and the cache serves any
request whose hash matches; behaviors whose hashes do not match (because their
prompts were edited) get fresh calls, which land as new events in the new run. In
\emph{strict} mode, behaviors re-fire and the runtime compares the live event
stream against the recorded one event by event; any divergence raises a replay
divergence error pinned to the first event that differs. A green strict replay is
a proof that the run is reproducible. This is also how the otherwise-unenforced
determinism contract is policed: a behavior that reads the wall clock will pass
permissive replay and fail strict replay, naming the offending event.

\section{Forking and structural diff}
\label{sec:forking}

The property that most distinguishes ActiveGraph from conventional agent
frameworks is the ability to ask, cheaply and honestly, ``what would have happened
if I had done this differently?'' A \emph{fork} branches a run at a chosen event.
The fork copies the parent's events up to and including the cutoff, then proceeds
with its own independent log; the two runs can be configured differently, run to
completion, and compared with a structural diff---all without disturbing the
parent.

\paragraph{Forks are cheap because the prefix is not re-executed.}
When a fork starts, the shared prefix is not recomputed by re-running behaviors.
The runtime replays the prefix against the fork's in-memory graph, serving every
model and tool response for those events from the content-addressed cache. No new
model calls, no new tool calls. Live execution resumes only at the cutoff. A fork
of a 200-step run that changes one setting at step 150 pays only for steps 150
onward; the first 149 steps, including their model calls, are free. Most agent
frameworks cannot fork at all, because their state is not reconstructable; those
that can typically must re-run the prefix, paying again for every model call.

\paragraph{Forks are honest because lineage is verifiable.}
A fork's relationship to its parent is not metadata that can drift out of sync; it
is the literal sharing of event ids up to the cutoff. The fork's events $1$ through
$k$ \emph{are} the parent's events $1$ through $k$. Anyone can verify the shared
lineage by reading the two logs. After the cutoff the fork has its own monotonic id
space, so ids do not collide. A structural diff between parent and fork then
reports exactly which objects, relations, and patches differ as a consequence of
the change introduced at the fork point.

\paragraph{Forks versus frames.}
The runtime offers a lighter-weight in-run primitive, the \emph{frame}, for
parallel sub-contexts that converge back within a single run and share that run's
log. The decision rule is durability: if the branches might diverge permanently or
need independent persistence, inspection, or diffing, fork; if they are
short-lived and reconverge, use a frame. A common pattern starts exploratory work
in frames and forks only the branches worth keeping.

\section{Worked example: a diligence pack}
\label{sec:worked}

The runtime ships with a reference \emph{pack}---a bundle of object types,
behaviors, tools, prompts, and policies for a domain. The diligence pack performs
investment due diligence: from a company name it generates research questions,
researches each against a document store, extracts claims with supporting
evidence, detects contradictions, identifies risks, and synthesizes a memo. It
ships with recorded fixtures, so the entire run executes offline, with no API key,
and is byte-deterministic.

The numbers below are reproducible by anyone. Running \texttt{activegraph}
\texttt{quickstart} (after \texttt{pip} \texttt{install} \texttt{activegraph}) executes the
bundled diligence demo on three companies (Northwind Robotics, Stellar Logistics,
Pinecone Bio) against recorded fixtures, requiring no API key and completing in
under thirty seconds.
Because all model and tool responses are served from fixtures, two runs produce
byte-identical logs; we re-ran the command and obtained the identical counts
reported here. (The repository's \texttt{examples/} directory contains several
other demos that exercise different subsets of the runtime---live-API runs, fork
walkthroughs, single-behavior claim extraction---and produce different, smaller
traces; the figures here are specifically those of the bundled \texttt{quickstart}
pack.)

The run produced \textbf{671 events}, yielding \textbf{93 objects} (3 companies, 24
questions, 9 documents, 25 claims, 25 evidence items, 1 contradiction, 3 risks, 3
memos) and \textbf{76 relations}, via \textbf{103 model calls} and 48 tool
calls---without any orchestration code. The behaviors fired reactively as objects
appeared on the graph.

\paragraph{Lineage is the deliverable.}
The point of the example is not the memo; it is that every statement in the memo is
traceable. A claim object such as \emph{``Northwind Q3 revenue grew 28\% YoY to
\$42M''} carries a provenance block naming the behavior that created it
(\texttt{document\_researcher}), the event that caused it, and the specific model
request event that produced it. It is linked by typed relations to the question it
\emph{addresses}, the document it is \emph{derived\_from}, and the evidence that
\emph{supports} it. For diligence, research, compliance, or scientific work---any
setting where the reasoning matters as much as the answer---this recoverable chain
from goal to output, reconstructable from the log alone, is the actual product.

\section{An architectural affordance for self-improving agents}
\label{sec:selfimprove}

We now state an implication of the architecture that we believe is significant but
that we explicitly do \emph{not} evaluate in this paper. We claim only that the
substrate removes specific obstacles to self-improvement; we do not present a
self-improving agent or measure one.

A self-improving agent modifies its own behavior---its rules, its prompts, its
tool repertoire. On a conventional architecture such self-modification is
dangerous precisely because it is invisible and irreversible: the agent changes
how it works, and there is no faithful record of the change and no way to undo it.
On ActiveGraph, rules and behaviors are part of the logged state, so a
self-modification is itself an event. This yields two affordances that
self-improvement specifically needs. First, \textbf{auditability and rollback}: a
run can be replayed as it was before a self-modification, and a strict replay will
reveal exactly how behavior diverged after it. Second, and more powerfully,
\textbf{fork-and-diff as an evaluation primitive}: the natural improvement loop is
to propose a change, fork the run at the point of proposal, apply the change in the
fork, run it forward, and structurally diff the outcome against the parent---and
because the shared prefix is served from cache, each candidate improvement is
evaluated without re-paying for the history that preceded it. The architecture thus
supplies the cheap, honest counterfactual comparison that any credible
self-improvement procedure requires. Turning these affordances into a working
self-improving system is future work.

\section{Related work}
\label{sec:related}

\paragraph{Agent memory systems.}
A growing body of work treats memory as a module layered onto an agent.
MemGPT/Letta~\cite{memgpt} introduces operating-system-style virtual memory
management, paging context in and out of a fixed window. Zep and its Graphiti
engine~\cite{zep} maintain a temporal knowledge graph as an internal memory
structure. Mem0~\cite{mem0} targets production memory extraction and retrieval. Closest to our
stance, Hindsight~\cite{hindsight} explicitly argues for memory as a ``first-class
substrate'' with epistemic separation of observation from inference and an
auditable update process---yet it remains a dedicated memory \emph{layer} feeding
an otherwise stateless model, rather than a substrate from which the entire run,
including rules and tool calls, is projected. These systems improve personalization and
context carry-over, but they share a stance: memory is derived state---summaries,
extractions, embeddings---layered onto an agent whose primary representation lives
elsewhere, and the provenance of what ends up in context is at best partial.
ActiveGraph takes the opposite stance: the log is primary and everything
else---including any memory view---is a projection of it, so provenance is total
and replay is exact. Our own prior research on graph-based agent memory motivated
this inversion: layered memory architectures repeatedly ran up against the absence
of a single authoritative history to project from.

\paragraph{Event sourcing and reactive dataflow.}
The substrate borrows from data-systems practice: event sourcing and CQRS, in which
state is a fold over an immutable event log, and incremental/reactive dataflow, in
which derived values recompute when their inputs change. ActiveGraph's contribution
is not these mechanisms but their application to agentic systems whose steps include
nondeterministic model calls, handled via the content-addressed response cache of
\S\ref{sec:replay}.

\paragraph{Blackboard architectures.}
The coordination model is older than agents. Blackboard systems~\cite{blackboard}
from the 1970s and 80s organized problem solving around a shared knowledge
structure that independent ``knowledge sources'' read from and wrote back to, with
no direct calls between them---structurally close to behaviors reacting to a shared
graph. Two things kept the model from scaling, and both were specific to their era.
The knowledge sources were brittle, narrow, hand-built expert-system components;
and the control logic deciding which source should act had to be authored by hand.
Large language models dissolve both constraints: a behavior can now be a general
LLM-backed routine rather than a narrow rule, and the coordinating logic can be
expressed in natural language or generated rather than hand-coded. We would go
further: the blackboard's shared-state, react-and-write-back model may fit LLMs
better than the conversational loop they are usually wrapped in. ActiveGraph is, in
this light, less a new idea than a vindication of an old one---with the addition
blackboard systems never had: an event-sourced substrate that makes the shared
state replayable, forkable, and fully traceable.

\paragraph{The BabyAGI lineage.}
ActiveGraph is a direct architectural successor to BabyAGI~\cite{babyagi}. Where
BabyAGI kept state in a global list mutated by a loop, ActiveGraph makes state a
projection of an append-only log mutated only through events, preserving the
original's task-generating, self-extending character while making every step
durable, inspectable, and replayable.

\section{Limitations and conclusion}
\label{sec:limitations}

\paragraph{Failure modes.}
A reactive substrate introduces failure modes a linear loop does not, and we name
them rather than imply the architecture is cleaner than real systems are. Because
behaviors emit events that trigger behaviors, a run can in principle diverge or
loop: the runtime's defense is a per-run budget (caps on events, behavior calls,
model calls, patches, recursion depth, wall-clock time, and cost), so a runaway
cascade halts against a ceiling rather than expanding unbounded---the budget is a
blunt instrument, not a static guarantee of termination. Replay cost grows with log
length, so very long-lived runs eventually need checkpointing or compaction, which
we do not yet provide; a million-event run is replayed in full today. Schema
evolution---changing an event or object type after events using the old shape are
already on disk---is handled by migration tooling but remains a real operational
burden. External tools with side effects are made replay-safe only by recording
their responses; a tool that mutates the outside world still mutates it on first
execution, and only the \emph{record} of that mutation replays deterministically.
Ordering within a single run is well-defined by the append-only log; concurrent or
distributed writers, and multi-agent contention over a shared graph, raise ordering
questions this paper does not resolve. These are the obvious places the model is
incomplete, and several are natural next work.

The approach has costs. The determinism contract places a real burden on behavior
authors and is enforced only dynamically, so violations surface at replay rather
than at write time. Determinism is achieved by recording model and tool responses,
which consumes store space proportional to run size and means that a fork which
changes a prompt must re-execute the affected calls. We report no large-scale
empirical evaluation of task performance; this is a systems and architecture
contribution, and the quantitative artifacts we present (event counts, provenance
chains, deterministic replay) are demonstrations of the mechanism rather than a
benchmark study. Establishing that the auditability and forking properties
translate into measurable advantages on downstream agent tasks---and building the
self-improvement loop sketched in \S\ref{sec:selfimprove}---is the natural next step.

What we have shown is that a single inversion---treating the append-only event log
as the agent rather than as its exhaust---buys a set of properties that are
otherwise hard to obtain together: deterministic replay over computations that
include model calls, counterfactual forks that do not re-pay for shared history,
and total lineage from goal to output. For the class of long-running agentic
systems where one must be able to explain, reproduce, and revise what the agent
did, these properties are not conveniences. They are the point.


\end{document}